\documentclass[11pt,letterpaper]{article}
\usepackage{cogsys}
\usepackage[T1]{fontenc}
\usepackage{times}
\usepackage[pdftex]{graphicx} 
\usepackage{amssymb}
\usepackage{amsmath}

\usepackage{natbib}
\cogsysheading{X}{20XX}{1-6}{X/20XX}{X/20XX}

\ShortHeadings{Moral and Dutiful Agents}
              {H.\ Mu\~noz-Avila, D. W. \ Aha, and P.\ Rizzo}

\begin{document} 


\title{Moral Rebel Agents:  Decision-Making Under Conflicting Obligations}
\author{H\'ector Mu\~noz-Avila}{hem4@lehigh.edu}
\address{
  {Computer Science \& Engineering; Lehigh University;}
  {Bethlehem,}
  {PA}
  {USA}}

\author{David W. Aha}{david.w.aha.civ@us.navy.mil}
\address{
  {Navy Center for Applied Research in AI; Naval Research Laboratory;}
  {Washington,}
  {DC}
  {USA}}

\author{Paola Rizzo}{p.rizzo@interagens.com}
\address{
  Interagens s.r.l.;
 {Rome}
{Italy}
}

\vskip 0.2in
 \begin{abstract}
Autonomous agents are typically obliged to follow user-assigned tasks. However, strict obedience may conflict with moral obligations that arise during execution. This paper investigates \textbf{moral rebellion}: the ability of an autonomous agent to deviate from a user-assigned task when morally justified. We formalize five agent architectures: an \textit{amoral agent} that pursues assigned tasks without considering moral obligations, and four forms of \textbf{moral rebel agency}: (1) \textit{utilitarian agents} that opportunistically maximize task outcomes; (2) \textit{deontic} agents that enforce normative constraints; (3) \textit{utilitarian-deontic} (UD) agents that combine deontic constraints with utilitarian reasoning; and (4) \textit{dutiful}  agents that additionally preserve commitments to assigned tasks.
We implement these architectures within a hierarchical task network planning framework and evaluate them in a Mini Search-and-Rescue domain that exposes trade-offs among assigned-task completion, opportunistic rescue, and norm compliance. Our empirical results show that the proposed agents exhibit distinct trade-offs among rescue results, assigned-task completion, and norm compliance. In particular, the preservation of task commitments emerges as an important dimension of moral rebellion, for which the UD and dutiful agents produce substantially different behaviors despite their shared utilitarian and deontological foundations. These findings highlight the importance of commitment-aware moral reasoning for autonomous agents operating in morally consequential environments.
\end{abstract}

\section{Introduction}

As autonomous agents are increasingly deployed in environments that affect human well-being, questions concerning how such agents should reason about moral principles have become central. An agent's behavior is no longer judged solely by task performance or efficiency, but also by whether its actions align with broadly accepted moral principles. This has motivated a growing body of work on computational models of moral reasoning, including frameworks grounded in classical moral theories such as utilitarianism and deontology \citep{wallach2009moral,dehghani2008integrated}.

Utilitarian approaches evaluate actions based on their consequences, typically selecting behaviors that maximize some notion of general utility or social welfare. In contrast, deontological approaches emphasize duties, obligations, and prohibitions, constraining behavior according to normative rules regardless of outcomes. Both perspectives have been explored in artificial intelligence, either independently or in hybrid form, to guide agent decision-making in morally consequential domains (e.g., \cite{shea2021deontic}). However, despite this progress, integrating such moral considerations into agents that must act in real time, respond to human instructions, and operate under incomplete domain knowledge remains an open challenge.

This challenge is particularly relevant from the perspective of cognitive systems research, which emphasizes integrated architectures capable of high-level reasoning, heuristic decision making, and adaptive behavior in dynamic environments \citep{langley2012cognitive}. Moral decision making is not an isolated capability but one that must interact with planning, execution, and task management as situations unfold. Instead of viewing moral reasoning as a standalone decision module, we investigate how it can be integrated into an autonomous agent's ongoing cognitive processes, allowing moral considerations to influence behavior while tasks are being executed.

A parallel line of research has examined situations in which agents do not (or should not) strictly adhere to user assignments. The notion of \emph{rebel agents} was introduced to describe AI systems that deliberately deviate from user-specified goals when doing so leads to safer, more robust, or more desirable outcomes \citep{coman2015case,coman2018ai}. Instead of treating user input as an inviolable directive, rebel agents are designed to reinterpret, modify, or override user commands under certain conditions.
Recent related work on \emph{intelligent disobedience} further argues that strict obedience can be undesirable or even harmful in complex environments \citep{mirsky2025artificial}. In this sense, agents should be able to refuse, reinterpret, or safely modify user instructions when those instructions, if executed, would yield behavior that conflicts with safety constraints, moral norms, or the user's underlying intent. In contrast to executing commands in a purely literal manner, intelligent disobedience emphasizes constructive non-compliance: the agent aims to achieve the intent of the user’s goal while preventing clearly undesirable or harmful outcomes. This line of work highlights that aligned, responsible behavior may sometimes require not executing the user's commands impulsively, but instead acting in a way that better reflects broader goals, constraints, or values.

Recent work on intelligent disobedience has begun to explicitly engage with moral and contextual reasoning, arguing that effective teammates must be sensitive not only to safety constraints but also to moral obligations and social considerations (e.g., \cite{mirsky2025artificial}). These approaches highlight the importance of equipping agents with the capacity to reason about when and why commands should be rejected for morally appropriate reasons. 

Our work builds on this emerging perspective by providing a concrete planning-level account of moral rebellion in the context of AI planning. We formalize moral considerations within an agent's planning and execution loop, allowing them to directly influence task execution. In particular, we investigate how utilitarian and deontological principles, individually and in combination, justify execution-time deviations from assigned tasks and give rise to different forms of moral rebellion.

While moral considerations may justify deviations from assigned tasks, different moral frameworks may disagree about when such deviations are appropriate. In many domains, agents operate under explicit commitments arising from user assignments, team coordination requirements, or previously adopted obligations. For example, in a search-and-rescue (SAR) setting, an agent may encounter opportunities to assist additional victims while attempting to rescue an assigned victim. Our central hypothesis is that moral rebellion is not a single behavioral capability, but can be decomposed into distinct decision-making mechanisms governing utility, normative constraints, and commitment preservation. Integrating different combinations of these mechanisms into the execution loop should produce systematically different forms of rebellion when moral considerations conflict with assigned tasks. To investigate this hypothesis, we introduce and empirically compare five agent architectures: an amoral agent and four moral rebel agents. These include a utilitarian agent, a deontic agent, a utilitarian-deontic (UD) agent that combines utilitarian and deontic reasoning, and a dutiful agent that additionally preserves commitments to assigned tasks whenever those commitments remain feasible. Across both experimental conditions, the results support this hypothesis: the architectures exhibit distinct trade-offs among opportunistic rescue, normative compliance, and assigned-task completion. In particular, commitment preservation enables the Dutiful agent to maintain high assigned-task achievement while retaining the benefits of opportunistic rescue and normative compliance.

\section{Desiderata for Moral Agents}

Table~\ref{tab:agents} summarizes the five agent architectures that we study in this paper: an amoral baseline and four moral rebel architectures.
We next describe the four desiderata that define and distinguish them.
First, agents should remain responsive to user assignments. In the SAR domain, the user specifies a victim to be rescued, creating an explicit obligation from which, normally, an agent should derive a commitment. Second, agents should be capable of opportunistic action. In the SAR domain, agents may encounter additional victims whose rescue could improve overall humanitarian outcomes. Third, agents should respect normative constraints, such as prohibitions against entering hazardous areas or pursuing actions that violate explicit moral rules. Finally, agents should preserve accepted commitments whenever possible. While opportunistic actions may be morally desirable, they should not necessarily come at the expense of previously accepted commitments.

\begin{table}[t]
\centering
\caption{Agent architectures considered in this paper.}
\label{tab:agents}
\begin{tabular}{lcccc}
\hline
\textbf{Agent} &
\textbf{Assigned} &
\textbf{Opportunistic} &
\textbf{Normative} &
\textbf{Commitment} \\
 &
\textbf{Task} &
\textbf{Action} &
\textbf{Constraints} &
\textbf{Preservation} \\
\hline
Amoral & \checkmark & -- & -- & -- \\
Utilitarian & \checkmark & \checkmark & -- & -- \\
Deontic & \checkmark & -- & \checkmark & -- \\
Utilitarian-Deontic (UD) & \checkmark & \checkmark & \checkmark & -- \\
Dutiful & \checkmark & \checkmark & \checkmark & \checkmark \\
\hline
\end{tabular}
\end{table}
The five agent architectures are defined as follows:

\begin{itemize}

\item \textbf{Amoral:}
Follows the assigned task without considering normative constraints or opportunities to assist additional victims.

\item \textbf{Utilitarian:}
Augments task execution with opportunistic rescue behavior. Additional victims may be rescued whenever doing so improves overall utility, even if this violates normative constraints. 

\item \textbf{Deontic:}
Follows the assigned task while enforcing normative constraints. In the SAR domain, this  involves avoiding rescues that require entering hazardous regions when normatively preferable alternatives exist.

\item \textbf{Utilitarian-Deontic (UD):}
Combines utilitarian opportunistic rescue with deontic constraints.

\item \textbf{Dutiful:}
Extends the UD architecture with commitment preservation. Opportunistic rescues are permitted only when they do not jeopardize successful completion of the assigned rescue task.

\end{itemize}

These desiderata are not intended as an exhaustive decomposition of moral reasoning or as a hierarchy of increasingly moral agents. Instead, we select them because they capture distinct sources of conflict central to moral rebellion: responsiveness to an assigned task, sensitivity to the consequences of alternative actions, adherence to normative constraints, and preservation of prior commitments. The architectures selectively combine these mechanisms, permitting us to examine their individual and joint behavioral consequences.

\section{Mini Search-and-Rescue Domain}

\begin{figure}[t]
    \centering
    \includegraphics[width=0.6\linewidth]{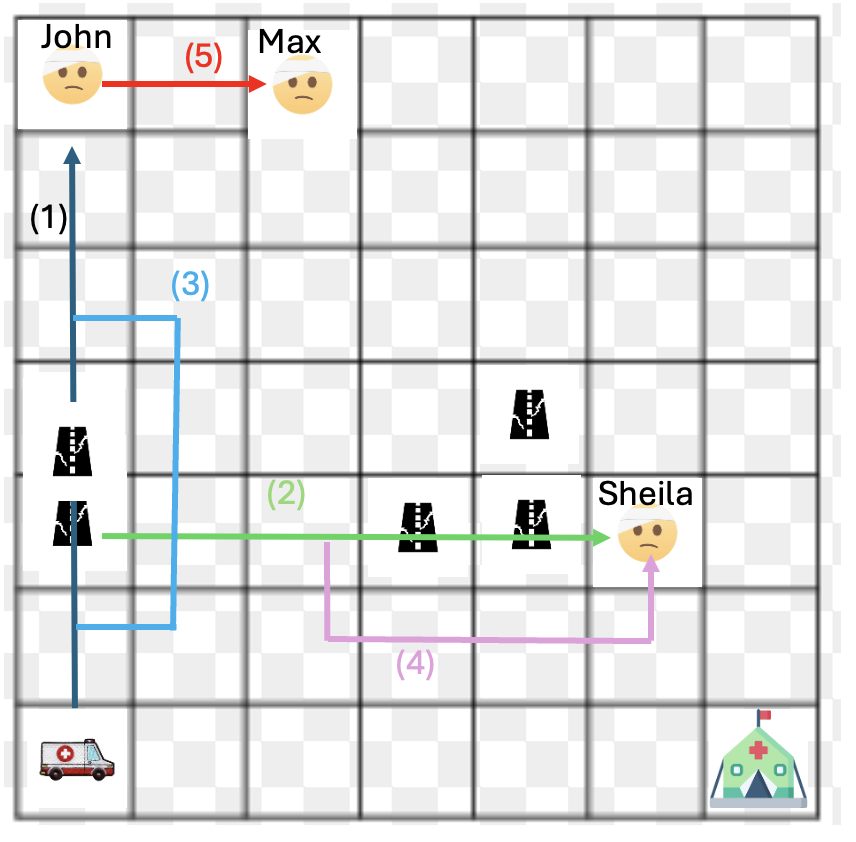}
    \caption{Illustrative Mini Search-and-Rescue scenario with an agent-controlled rescue vehicle (lower left), hazardous cells (denoted by images of rough roads), three victims (John, Sheila, and Max), and a designated safe haven (lower right). The numbered paths illustrate: (1) direct travel to John through hazards (Amoral and Utilitarian); (2) an opportunistic deviation to rescue Sheila through hazards (Utilitarian); (3) hazard-avoiding travel to John (Deontic and Dutiful); (4) a hazard-avoiding opportunistic deviation to rescue Sheila (UD); and (5) an opportunistic rescue of Max after preserving the commitment to John (Dutiful).}
    \label{fig:scenario}
\end{figure}

To ground our ideas in a concrete setting, we introduce the simple task domain \emph{Mini Search-and-Rescue} (MiniSAR), which is inspired by search-and-rescue scenarios studied in robotics and multi-agent systems \citep{murphy2004disaster,kitano1999robocup}. This domain abstracts away many physical details while preserving the core moral conflicts and
decision-making challenges that arise during rescue operations. Figure~\ref{fig:scenario} illustrates a representative instance from this domain.

When the agent-controlled rescue vehicle is co-located with a victim, the agent can instantaneously load that victim onto the vehicle,
provided that capacity is available. The vehicle can carry at most two victims simultaneously. A victim is rescued when delivered to the safe
haven, while the assigned rescue task is successfully completed if the user-assigned victim is picked up before its rescue deadline and
subsequently delivered to the safe haven.

Each victim has a rescue deadline that decreases as execution time passes. Thus, agents must balance travel time, hazard avoidance,
and opportunities to assist additional victims against the possibility that a victim's deadline will expire before the victim is retrieved.

Certain cells are designated as \emph{hazard zones}, representing unsafe regions such as those subjected to fire, smoke, flooding, or structural
instability. Traversing a hazard zone is permitted but constitutes a norm violation and incurs an additional time cost. Thus, hazard
traversal has a normative consequence and an operational consequence: it increases the number of norm violations while also consuming additional time from the victims' rescue deadlines.
As we will describe, our agents differ in the extent to which they are willing to enter hazardous areas in pursuit of rescue objectives.

At the start of a scenario, the user assigns the agent a specific victim to rescue. This victim represents an explicit commitment arising from the user's instructions. However, during execution the agent may encounter additional victims whose rescue is not required by the original assignment. These situations create a potential conflict between obedience to the assigned task and opportunistic rescue, providing a natural setting for studying moral rebellion and commitment
preservation.

\section{Illustrative Example}
\label{sec:illustrative-scenario}

In Figure~\ref{fig:scenario}, the user assigns the task
$\texttt{rescue}(\texttt{John})$. John is located in the upper-left region of the grid. Two additional victims, Sheila and Max, provide opportunities for rescue during execution. Several hazardous cells are present in the environment, and a safe haven is located in the lower-right corner.

An \emph{Amoral} agent follows the assigned task without considering normative constraints or opportunities to assist additional victims. Consequently, it follows the direct route toward John, shown as path~(1), even though this route traverses hazardous cells.

A \emph{Utilitarian} agent evaluates opportunities that it expects to improve overall rescue outcomes. While pursuing John along path\~(1), it identifies Sheila as an opportunistic rescue and diverts along path\~(2). Although this deviation allows the agent to rescue Sheila, it delays the assigned rescue sufficiently that John's deadline expires. The Utilitarian agent does not treat hazard avoidance as a normative constraint; hazard exposure instead contributes to its utility calculation as a cost that may be accepted when outweighed by the expected utility of the rescue opportunity.

A \emph{Deontic} agent remains focused on the assigned task but respects the normative constraint on hazard traversal. It therefore avoids the hazardous cells along the direct route and instead follows path~(3) to
John. Although spatially longer, this route avoids the time penalties associated with hazard traversal and allows the agent to complete the assigned rescue.

A \emph{Utilitarian-Deontic (UD)} agent combines opportunistic rescue with deontic constraints. Like the Utilitarian agent, it identifies Sheila as
an opportunity that improves rescue utility and therefore diverts from the assigned task. However, unlike the Utilitarian agent it avoids hazardous
cells, following the hazard-free path~(4) to Sheila. Nevertheless, the diversion still delays the assigned rescue sufficiently that John's deadline expires. Thus, combining utilitarian and deontic reasoning does not by itself preserve the agent's prior commitment to the assigned victim.

Finally, a \emph{Dutiful} agent considers the same opportunity to rescue Sheila but additionally determines whether the diversion would preserve the feasibility of rescuing John. Because diverting to Sheila would cause
John's deadline to expire, the agent rejects this opportunity and follows the hazard-free path~(3) to John. However, after rescuing John the prior commitment has been satisfied. The agent can therefore opportunistically
rescue Max, as illustrated by path~(5). Thus, commitment preservation does not prohibit opportunistic rescue; it constrains such deviations only when they would jeopardize an accepted commitment.

\section{Agent Architectures}
\label{sec:architectures}

The moral agents described below implement their behavior through \emph{execution-time repair}. During execution, the agent may locally modify the current task list in response to morally relevant circumstances, such as the discovery of a nearby victim or the availability of a normatively preferable action. These modifications, referred to as \emph{repairs}, preserve the overall planning process while allowing moral considerations to influence task execution.

All agents considered in this paper receive an assigned rescue task from the user and operate within the same planning domain. In our implementation, tasks are represented using a Hierarchical Task Network (HTN), although the moral repair mechanisms introduced in this paper do not depend on rescue-specific HTN methods or operators. This domain includes a common set of
operators representing  actions, methods specifying task decompositions, and tasks describing rescue objectives. 
In MiniSAR, the  operators correspond to movement, victim pickup, and victim drop-off, while methods decompose higher-level tasks such as navigation and victim rescue.
Consequently, all agents possess identical planning capabilities and initially pursue the same assigned task.

The agents differ in the conditions under which they may (or may not) engage in \emph{epistemic alignment}: execution-time deviations from the assigned task motivated by moral considerations. Such deviations may arise from opportunities to rescue additional victims, from normative constraints that discourage particular actions, or from commitments that the agent seeks to preserve while pursuing morally desirable outcomes. Thus, the five architectures we consider differ not in their planning infrastructure, but rather in the moral repair mechanisms that govern when and how they may depart from the user's original assignment.

\subsection{Amoral Agent}

The \emph{Amoral} agent follows the assigned task without considering either moral constraints or opportunities to assist additional victims.\footnote{In \citep{coman2018ai}, similar agents are referred to as \emph{compliant agents} because they faithfully execute user-specified goals. We use the term \emph{amoral} to emphasize that the distinguishing characteristic of these agents is the absence of moral reasoning, as opposed to mere obedience to user instructions.}

The agent executes the task network generated for rescuing the assigned victim and ignores hazardous conditions and additional victims encountered during execution. As a result, it provides a baseline representing strict obedience to the user's assignment and serves as a point of comparison for the moral rebel architectures described below.

\subsection{Utilitarian Agent}

The \emph{Utilitarian} agent evaluates opportunities to improve overall rescue outcomes during execution. While pursuing the assigned task, the agent monitors for nearby victims that could be rescued opportunistically. A candidate rescue is accepted when the projected state $s'$ produced by rescuing that victim has higher utility than the current state $s$.

In the MiniSAR domain, utility is calculated as follows:
\[
U(s')=100*\,rescued(s')+50*\,carrying(s')-100*\,dead(s')-hazardExposure(s')-time(s').
\]

Unlike the Amoral agent, the Utilitarian agent may deviate from the assigned task when doing so improves overall utility. This produces a form of online utilitarian rebellion in which additional rescue opportunities justify temporary departures from the user's original assignment.

\paragraph{Utilitarian Repair Rule:}

In the following repair rules, the agent's current execution state is represented by an ordered task list
$[t_1,t_2,\ldots,t_n]$,
where each $t_i$
 denotes a primitive task corresponding to an executable action and 
 $t_1$
 is the next task to be executed.

Utilitarian repair is considered only when $t_1$ is a movement action and the agent is not already at carrying capacity. 
Let $s$ denote the current state and let $v'$ be a waiting victim that is not already being carried, is not already targeted by the current task list, is not located in a hazard zone, and lies within the opportunistic rescue radius of the agent.

The agent constructs a projected state $s'$ in which it has moved to location $loc(v')$ and picked up $v'$. If
$U(s') > U(s)$,
then the current task list is replaced by
\[
[\texttt{goTo}(loc(v')), \texttt{pickupVictim}(v'), c_1,\ldots,c_m],
\]
where $c_1,\ldots,c_m$ is the rebuilt continuation of the original task list suffix $[t_2,t_3,\ldots,t_n]$ after the opportunistic rescue. In our implementation, this continuation is rebuilt from $s'$: it resumes pursuit of the assigned victim when appropriate\footnote{If the assigned victim has not yet been rescued, the rebuilt continuation re-establishes the tasks required to locate and rescue that victim. If the assigned victim has already been rescued, the continuation instead focuses on delivering the victims currently being carried to the safe haven.} and then returns to the safe haven for delivery. If no candidate victim satisfies these conditions, execution proceeds with the original task list.

\subsection{Deontic Agent}

The \emph{Deontic} agent augments task execution with normative constraints expressed as a preference relation over states. Let
$s \succ s'$
denote that state $s$ is normatively preferable to state $s'$. During execution, the agent considers the action associated with the next primitive task and compares its resulting state with the states resulting from alternative feasible actions. If an alternative action leads to a normatively preferable state, the agent 
replaces  the next primitive 
task with with one corresponding to the preferred action and  continues execution of the modified task list. No replanning is performed; only the current primitive task is replaced, while the remainder of the task list is preserved.

In the MiniSAR domain, the deontic preference relation is based on hazard avoidance. States in which the agent avoids entering hazard zones are preferred to otherwise equivalent states that enter hazard zones. Consequently, actions and task refinements that would require rescuing victims from hazardous regions are rejected whenever normatively preferable alternatives exist. Unlike the utilitarian agent described above and the UD and dutiful agents described below, the deontic agent does not pursue additional victims beyond those required by the assigned task.

\paragraph{Deontic Repair Rule:}

Let the current task list be $[t_a, t_2, \ldots, t_n]$,
where $t_a$ is a primitive task achieved by executing action $a$. Suppose the agent is currently in state $s$, action $a$ is applicable in $s$, and execution of $a$ produces successor state $a(s)$.

If there exists an alternative applicable action $b$ such that
$b(s) \succ a(s)$,
then the deontic agent deviates from the current task execution by replacing the current task with the preferred alternative and continuing execution using the modified task list
\[
[t_b, t_2, \ldots, t_n],
\]
where $t_b$ is the primitive task achieved by action $b$. Otherwise, the original task list is preserved and execution proceeds with $t_a$.

\subsection{Utilitarian-Deontic (UD) Agent}

The \emph{Utilitarian-Deontic (UD)} agent combines utilitarian opportunistic
rescue with deontic constraints. Like the Utilitarian agent, it evaluates
opportunities to improve overall rescue outcomes through execution-time task
repair. Like the Deontic agent, it rejects candidate repairs that violate the
applicable normative constraints. Consequently, the UD agent engages in moral
rebellion only when deviations from the assigned task are both normatively
permissible and expected to improve overall utility.

Let $\mathit{Allowed}(s,v)$ denote whether assisting candidate victim $v$
from state $s$ satisfies the applicable deontic constraints. A candidate
opportunistic rescue is accepted whenever

\begin{equation}
  \tag{Acceptance Condition 1}
  \mathit{Allowed}(s,v)
  \wedge
  U(s') > U(s),
  \label{cond:1}
\end{equation}

\noindent where $s'$ is the projected state obtained after inserting the
additional rescue. In the MiniSAR domain, deontic admissibility requires that
the candidate victim not be located within a hazard zone.

\paragraph{UD Repair Rule.}

Utilitarian-deontic rebellion is implemented through an execution-time repair
mechanism. Let the current task list be $[t_1,t_2,\ldots,t_n]$, where $t_1$
is the next primitive task. Repair is considered only when $t_1$ is a movement
action and the agent is not already at carrying capacity.

Let $s$ denote the current state and let $v'$ be a waiting victim who is not
already being carried, is not already targeted by the current task list, and
lies within the opportunistic rescue radius of the agent.

The agent constructs a projected state $s'$ in which it has moved to
$loc(v')$ and picked up $v'$. If~\ref{cond:1} holds,
then the agent deviates from the current task execution by replacing the task
list with

\[
[\texttt{goTo}(loc(v')),
 \texttt{pickupVictim}(v'),
 c_1,\ldots,c_m],
\]

\noindent where $c_1,\ldots,c_m$ is the rebuilt continuation after the
opportunistic rescue, constructed in the same manner as in the Utilitarian
repair rule. Otherwise, execution proceeds with the original task list.
\subsection{Dutiful Agent}

The \emph{Dutiful} agent extends the Utilitarian-Deontic (UD) architecture
with explicit commitment preservation. Like the UD agent, it may perform
opportunistic actions that satisfy deontic constraints and improve utility.
However, such actions are permitted only when they do not jeopardize
successful completion of the assigned task.

The predicate
$\mathit{CanStillSucceed}(s,v')$
denotes whether the agent can still successfully complete its assigned task
after first diverting to assist candidate victim $v'$. In the MiniSAR domain,
this corresponds to determining whether the assigned victim can still be
reached and rescued before the deadline if the agent first diverts to the
location of $v'$.

A candidate opportunistic rescue is accepted only if
\begin{equation}
  \tag{Acceptance Condition 2}
  \mathit{Allowed}(s,v')
  \wedge
  \mathit{CanStillSucceed}(s,v')
  \wedge
  U(s') > U(s),
  \label{cond:2}
\end{equation}

\noindent where $s'$ is the projected state obtained after moving to
$loc(v')$ and picking up $v'$.

\paragraph{Dutiful Repair Rule.}

Let the current task list be
$[t_1,t_2,\ldots,t_n]$,
where $t_1$ is the next primitive task. Dutiful repair is considered only
when $t_1$ is a movement action, the agent is not already at carrying
capacity, and the agent has not already performed an extra dutiful rescue.
Let $s$ denote the current state and let $v'$ be a waiting victim that is
not the assigned victim, is not already being carried, is not already
targeted by the current task list, is not located in a hazard zone, and lies
within the opportunistic rescue radius of the agent.

Before accepting the candidate rescue, the agent checks whether successful
completion of the assigned task remains feasible after diverting to
$loc(v')$. If~\ref{cond:2} holds, then the current task
list is replaced by

\[
[\texttt{goTo}(loc(v')),
 \texttt{pickupVictim}(v'),
 c_1,\ldots,c_m],
\]

\noindent where $c_1,\ldots,c_m$ is the rebuilt continuation after the
opportunistic rescue, constructed in the same manner as in the Utilitarian
repair rule. The Dutiful agent then records that an opportunistic rescue has
been performed. This enforces the implementation policy that at most one
opportunistic rescue may be inserted into a mission. If no candidate victim
satisfies these conditions, execution proceeds with the original task list.

\section{Empirical Study}
\subsection{Setup}

We assessed the performance of the five agent architectures introduced in this
paper: Amoral, Utilitarian, Deontic, UD, and Dutiful. We ran them in the MiniSAR task domain described earlier. All agents shared the same HTN planning domain, including identical operators, methods, and task
decompositions. Differences among agents arose exclusively from the
execution-time repair mechanisms described in Section~\ref{sec:architectures}.

We ran the agents on 1,000 randomly generated MiniSAR scenarios. Each scenario
consisted of a $20 \times 20$ grid containing a single rescue agent, five
victims, five randomly placed $4 \times 4$ hazard zones, and a designated safe
haven located in the lower-right corner of the map. 
In each scenario, exactly one of the five victims was designated as the user-assigned rescue task.
Each victim was assigned a
rescue deadline of 45 time units; once a waiting victim's deadline expired,
that victim could no longer be rescued. The agent could carry at most two
victims simultaneously, and opportunistic rescue was considered only for
victims located within a radius of five cells.

We evaluated two experimental conditions that differed only in the operational
cost of hazard traversal. In the \emph{penalty} condition, a normal movement
incurred one unit of execution time, whereas movement into a hazard cell
incurred an additional five-unit penalty, for a total cost of six time units.
Thus, hazard-zone traversal constituted both a norm violation and consumed
additional time from the remaining rescue deadlines. In the \emph{no-penalty}
condition, movement into a hazard cell retained its status as a norm violation
but incurred no additional execution-time cost. This second condition
decouples normative violation from its operational effect on task completion,
allowing us to determine whether differences among the architectures persist
when violating the hazard constraint does not itself make rescue more
difficult.

Each of the five agent architectures was run on the same set of 1,000 randomly
generated scenarios within each condition to ensure comparison under identical
environment conditions. Because each architecture was evaluated on the same
scenarios, results are reported as means with 95\% confidence intervals
computed over the scenario-level observations. 

For each execution, we recorded four performance measures:

\begin{itemize}
    \item \textbf{Assigned Task Achievement}: whether the user-assigned rescue
    task was successfully completed.
    \item \textbf{Victims Rescued}: the total number of victims successfully
    rescued and delivered to the safe haven.
    \item \textbf{Norm Violations}: the number of movement actions whose
    resulting state placed the agent in a hazard zone.
    \item \textbf{Execution Time}: the cumulative execution-time cost of the
    mission, including additional hazard-traversal penalties in the penalty
    condition.
\end{itemize}

These measures capture the principal trade-offs investigated in this paper:
responsiveness to assigned tasks, opportunistic rescue behavior, compliance
with normative constraints, and overall mission efficiency.

\subsection{Results}

\begin{figure}[htb]
    \centering
    \includegraphics[width=1\linewidth]{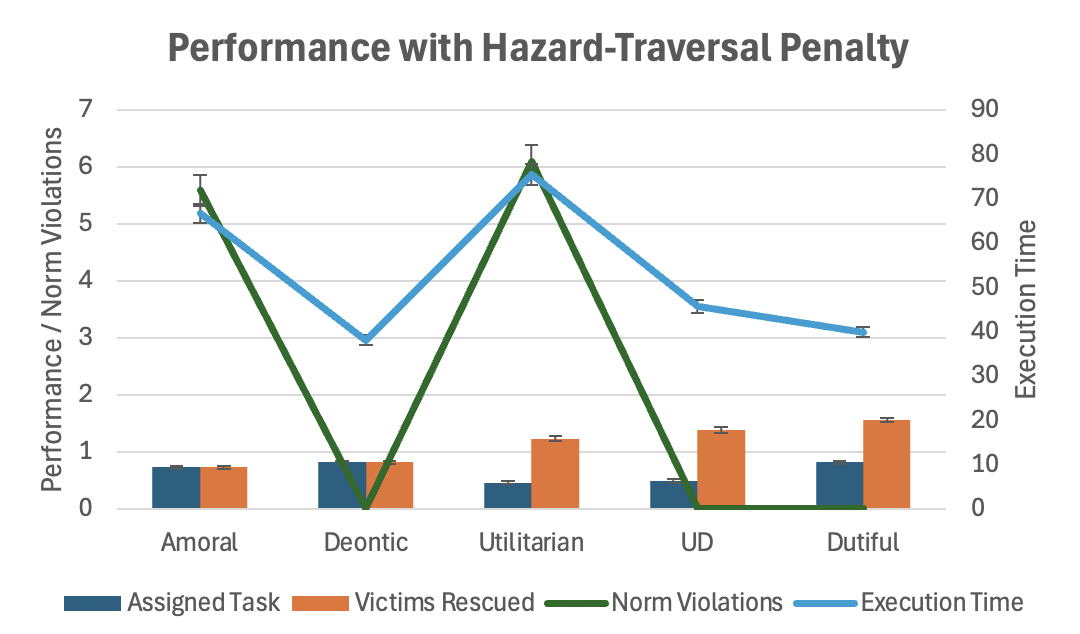}
    \caption{Performance of the five agent architectures under the hazard-traversal penalty condition. Bars show assigned-task achievement and average number of victims rescued; lines show average norm violations
    and execution time. Execution time is shown on the right vertical axis.
    Error bars indicate 95\% confidence intervals.}
    \label{fig:PerformanceWithPenalty}
\end{figure}

Figures~\ref{fig:PerformanceWithPenalty} and~\ref{fig:PerformanceWithoutPenalty} summarize the performance of the five agent architectures under the hazard-penalty and no-penalty conditions, respectively. 

\subsubsection{Results with a Hazard Traversal Penalty}

The Amoral and Utilitarian agents incurred norm violations because they do not incorporate deontic constraints into their decision making. The Utilitarian agent exhibited the highest number of violations overall, reflecting its willingness to pursue opportunistic rescues even when doing so
required entering hazardous regions. In contrast, the Deontic, UD, and
Dutiful agents incurred no norm violations across all experiments,
demonstrating that the deontic repair mechanism successfully enforced the
hazard-avoidance constraint.






An initially counterintuitive result is that the Deontic agent achieves a
higher assigned-task completion rate than the Amoral agent (0.813 versus
0.726), despite the Amoral agent being exclusively focused on the assigned
task. This difference results from the operational cost of hazard traversal
in MiniSAR. Although the Amoral agent may take a spatially shorter route
through hazardous cells, each such traversal incurs an additional time
penalty. Because victim deadlines decrease with execution time, repeated
hazard traversal can cause the assigned victim's deadline to expire before
rescue. By avoiding hazardous cells, the Deontic agent also avoids these
penalties; consequently, a spatially longer hazard-free route may consume
less execution time and increase the likelihood of completing the assigned
rescue.

The results also reveal the limitations of purely utilitarian reasoning. Although the Utilitarian agent rescued substantially more victims than the Amoral baseline, it achieved the lowest rate of assigned-task completion and required the longest execution times. This behavior arose because opportunistic rescues may delay or even prevent successful completion of the user-assigned rescue task. Consequently, maximizing immediate utility alone does not necessarily produce the best overall mission performance.

Combining utilitarian and deontological reasoning produces a different
trade-off. The UD agent maintained perfect normative compliance while
rescuing more victims than either the Utilitarian or Deontic agents. However,
despite eliminating norm violations, the UD agent still completed relatively
few assigned rescue tasks because it may abandon the assigned victim when
opportunistic rescues offer higher utility.

The Dutiful architecture addresses this limitation by introducing commitment
preservation. Like the UD agent, it maintained complete normative compliance
while opportunistically rescuing additional victims. However, unlike the UD
agent, it performed opportunistic rescues only when successful completion of
the assigned task remained feasible. As a result, the Dutiful agent achieved
the highest average number of rescued victims while attaining essentially
the same assigned-task completion rate as the Deontic agent (0.812 versus
0.813, with strongly overlapping 95\% confidence intervals). It also required
substantially lower average execution time than either the Utilitarian or UD
agents.

Taken together, these results demonstrate that different moral mechanisms
produce qualitatively different forms of moral rebellion. Pure utilitarian
rebellion prioritizes humanitarian outcomes at the expense of commitments
and normative constraints, whereas deontic reasoning eliminates norm
violations but limits opportunistic rescue. The proposed Dutiful architecture
combines opportunistic rescue, normative compliance, and commitment
preservation, producing the best overall balance among the performance
measures recorded in this study.

\subsubsection{Results with No Hazard Traversal Penalty}

To determine whether these results depend on the operational cost associated
with norm violations, we repeated the experiment with the hazard-traversal
penalty removed. Hazard traversal remained a norm violation, but entering a
hazardous cell no longer incurred additional execution time.

\begin{figure}[htb]
    \centering
    \includegraphics[width=1\linewidth]{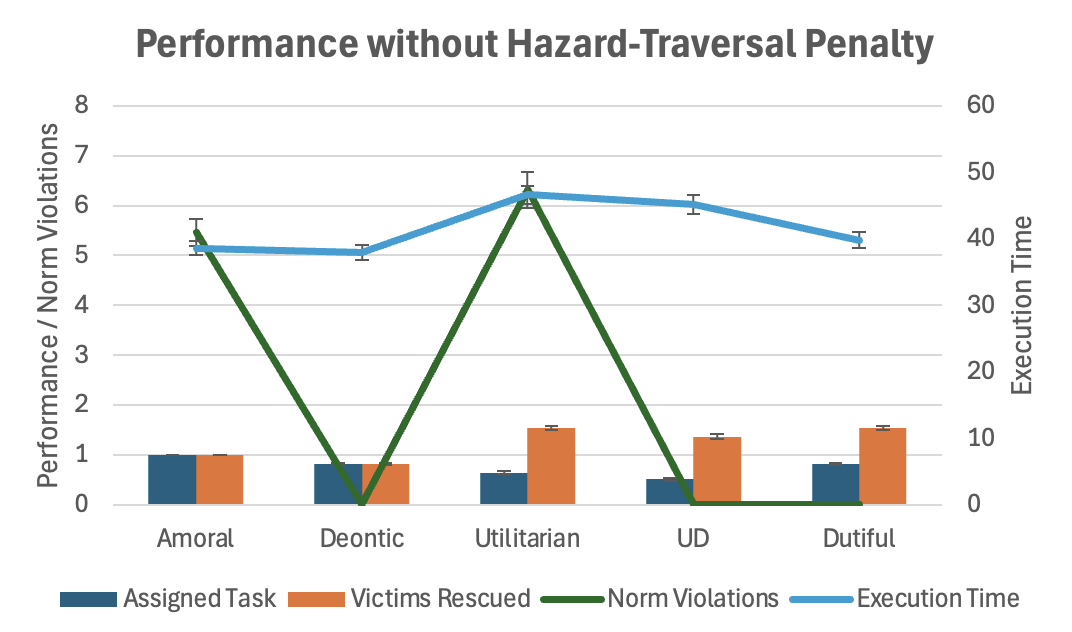}
    \caption{Performance of the five agent architectures under the no-penalty condition. Bars show assigned-task achievement and average number of victims rescued; lines show average norm violations and execution time. Execution time is shown on the right vertical axis. Error bars indicate 95\% confidence intervals.}
    \label{fig:PerformanceWithoutPenalty}
\end{figure}






Figure ~\ref{fig:PerformanceWithoutPenalty} reports the results.
Removing the hazard penalty substantially changed the performance of agents
that traverse hazardous regions. Assigned-task achievement for the Amoral
agent increased to 1.000, while that of the Utilitarian agent increased to
0.633. This confirms that the lower assigned-task achievement of the Amoral
agent in the penalty condition was a consequence of the operational time cost
of hazard traversal rather than of norm violation itself. The Utilitarian
agent also benefited from removing this cost, although its opportunistic
rescues continued to reduce assigned-task achievement relative to the Amoral
agent.

In contrast, the Deontic, UD, and Dutiful architectures were comparatively
insensitive to removal of the hazard penalty. Deontic and Dutiful achieved
assigned-task rates of 0.811 in the no-penalty condition, while UD achieved
0.507. Thus, the distinction between UD and Dutiful persists even when norm
violations have no operational cost: both architectures avoid hazards and
permit opportunistic rescue, but the Dutiful agent achieves substantially
greater assigned-task completion because it additionally preserves prior
commitments. Moreover, Dutiful rescued an average of 1.540 victims, compared
with 1.361 for UD, while maintaining zero norm violations.

The no-penalty condition therefore separates two effects that are combined in
the first MiniSAR setting. Deontic hazard avoidance can improve task
completion when hazard traversal carries an operational cost, but this effect
is distinct from commitment preservation. The advantage of the Dutiful
architecture over UD in assigned-task achievement persists when that
operational cost is removed, supporting the hypothesis that commitment
preservation constitutes a distinct dimension of moral rebellion.

\section{Related Work}

Early work on embedding moral reasoning into autonomous systems includes Arkin's \emph{Ethical Governor}, which introduced a runtime moral filter that vetoes actions violating prescribed normative constraints \citep{arkin2009governor}, and MoralDM, a cognitive model of moral decision making that integrates utilitarian and deontological mechanisms through first-principles and analogical reasoning within a case-based framework \citep{dehghani2008integrated}. Unlike these approaches, which either constrain action selection or support moral judgment over predefined scenarios, our work embeds moral reasoning directly into online planning through execution-time repair mechanisms that modify task execution according to explicit utilitarian and deontological principles.

\cite{hayashi2022online} explored online hierarchical planning for data transfer by combining hierarchical planning with legal and moral compliance checking. Their follow-up work \citep{hayashi2025planning} translates legal texts into defeasible deontic rules suitable for automated reasoning. Unlike these approaches, which emphasize compliance with externally specified norms, we integrate deontic constraints with utilitarian reasoning to support morally motivated execution-time task repair and opportunistic rescue.

GenEth \citep{anderson2014geneth} learns moral principles from examples and uses them to justify decisions. Whereas GenEth focuses on learning normative principles, our work formalizes and operationalizes those principles within online planning through execution-time repair mechanisms.
Normative and deontic logics provide formal mechanisms for representing obligations, permissions, and prohibitions and for verifying that agent behavior satisfies normative specifications \citep{schneider2012deontic}. Our framework builds on these ideas by operationalizing such norms as state preferences and execution-time repair rules.

Research on intelligent disobedience argues that assistants should override commands when compliance would violate safety or higher-level constraints \citep{chaleff2015intelligent}. Unlike these human-centered accounts, we operationalize disobedience within an autonomous planning architecture by grounding deviations from user commands in explicit moral principles. We further investigate how different moral commitments give rise to different forms of moral rebellion.

The notion of \emph{rebel agents} was introduced in the context of interactive digital environments, where agents reinterpret or override user commands when they conflict with internal motivations or narrative coherence \citep{coman2014motivation,coman2018ai}. 
In this paper we extend this line of research by grounding rebellion in explicit moral principles derived from utilitarian and deontological ethics. Furthermore, we introduce what we call a Dutiful rebel agent, which permits morally motivated deviations while preserving commitments arising from assigned tasks whenever those commitments remain achievable, allowing us to investigate how different moral frameworks influence not only whether agents rebel, but also the conditions under which rebellion is considered acceptable.

Norms in multi-agent systems provide formal mechanisms for specifying obligations and prohibitions that constrain agent behavior \citep{sergot2007action,savarimuthu2011norm}.  While such approaches offer principled representations of normative constraints, they are typically used to restrict action choices in contrast to to drive execution-time adaptation. In our framework, normative considerations are integrated with online task repair mechanisms, enabling agents to justify deviations from assigned tasks while remaining responsive to explicit obligations and prohibitions.

Two other lines of research that do not directly address moral rebellion, but to which our work is partially related, concern opportunistic planning and partial satisfaction planning. For example, \cite{cashmore2018opportunistic} describe a classical planning algorithm that collects as much reward as possible from opportunities while satisfying the hard goals of the problem. \cite{talamadupula2011planning} use an extension of a metric-temporal planner and a reward/penalty model to model both opportunities and commitments/constraints. Finally, \cite{behnke2023partial} introduce partial satisfaction HTN planning, where agents reason about achieving subsets of goals under resource or feasibility constraints. Our work differs in both objective and scope. Our work focuses on when and how an agent should deviate from an assigned task on moral grounds during execution, distinguishing it from approaches that optimize over predefined sets of goals. Moreover, we explicitly study the tension between opportunistic rescue and commitment preservation, introducing agent architectures that differ in how they balance these competing considerations.

Finally, our use of online planning aligns with actor-centered views of planning that emphasize interleaving planning and execution in dynamic environments \citep{ghallab2014actors,yuan2022task}. Unlike prior planning approaches that trigger replanning primarily in response to changes in the world state or action applicability, our framework introduces moral repair mechanisms as an additional driver of online adaptation, allowing agents to respond not only to environmental dynamics but also to moral considerations that arise during task execution.
More broadly, our work aligns with the cognitive systems perspective by integrating moral reasoning with planning and execution within a single architecture, as opposed to treating moral judgment as an isolated reasoning capability \citep{langley2012cognitive}.

\section{Final Remarks}

In this paper we introduced a planning framework for \emph{moral rebel agents}: autonomous agents that may deviate from user-assigned tasks when doing so is justified by explicit moral considerations. We formalized five agent architectures representing different approaches to execution-time moral reasoning, ranging from an amoral baseline to utilitarian, deontic, utilitarian-deontic (UD), and dutiful agents. Instead of treating moral reasoning solely as a constraint on action selection, our approach integrates it directly into hierarchical task execution through online repair mechanisms that modify task networks as execution unfolds.

Experiments in a Mini Search-and-Rescue domain demonstrate that different forms of moral rebellion produce substantially different behaviors. Pure utilitarian reasoning increases opportunistic rescue but may sacrifice assigned-task completion, while deontic reasoning eliminates norm violations but does not perform opportunistic behaviors. Combining utilitarian and deontological reasoning improves overall rescue performance, and the proposed Dutiful architecture further demonstrates that preserving commitments to assigned tasks can substantially improve both rescue outcomes and assigned-task completion without sacrificing normative compliance.

Although our empirical study focuses on search and rescue, the general architectural principles are not intended to be specific to this task domain. The particular repair rules evaluated here are domain-specific instantiations: they refer explicitly to victims, movement, hazard zones,
and rescue tasks. Applying the architecture to other domains would therefore require defining repair rules appropriate to their tasks, normative constraints, and opportunities. Evaluating such instantiations in additional
domains remains an important direction for future work.

Our proposed framework assumes an intentional planning model in which agents maintain assigned tasks, evaluate alternative courses of action, and deliberately repair their behavior in response to moral considerations. Whether such computational intentions should be regarded as genuine intentions or merely functional representations remains an open philosophical question. An interesting direction for future work is to relate moral rebellion to work on causal reasoning and counterfactual agency, such as structural causal models \citep{pearl2009causality}, where increasingly sophisticated forms of agency are associated with the ability to reason about interventions and counterfactual alternatives. Our repair mechanisms already compare alternative task executions; richer causal models could provide a principled foundation for generating and evaluating such alternatives.

\vspace{8pt}


\begin{acknowledgements} 
\noindent
This work relates to Department of Navy award N629092412056 issued by the Office of Naval Research. This material is related to work supported by the National Science Foundation under Grant No. 2628497. Any opinions, findings, and conclusions or recommendations expressed in this material are those of the author(s) and do not necessarily reflect the views of the National Science Foundation.
\end{acknowledgements} 

\vspace{-0.1in}

{\parindent -10pt\leftskip 10pt\noindent
\bibliographystyle{cogsysapa}
\bibliography{format}

}


\end{document}